\documentclass[english]{article}

\usepackage {algpseudocode}
\usepackage {algorithm}
\usepackage{longtable}
\usepackage{graphicx}

\usepackage{amssymb}

\begin{document}




\title{Improving Classification Neural Networks by using Absolute activation function (MNIST/LeNET-5 example)}


\author{Oleg I.Berngardt}

\maketitle

\begin{abstract}
The paper discusses the use of the Absolute activation function in
classification neural networks. An examples are shown of using this
activation function in simple and more complex problems. Using
as a baseline LeNet-5 network for solving the MNIST problem,
the efficiency of Absolute activation function is shown in comparison
with the use of Tanh, ReLU and SeLU activations. 
It is shown that in deep networks Absolute activation does not cause 
vanishing and exploding gradients, and therefore
Absolute activation can be used in both simple and deep neural networks. 
Due to high volatility of training networks with Absolute activation, 
a special modification of ADAM training algorithm is used, that estimates 
lower bound of accuracy at any test dataset using validation dataset analysis
at each training epoch, 
and uses this value to stop/decrease learning rate, and reinitializes 
ADAM algorithm between these steps.
It is shown
that solving the MNIST problem with the LeNet-like architectures based on Absolute activation 
allows to significantly reduce the number of trained parameters 
in the neural network with improving the prediction accuracy. 
\end{abstract}











\section{Introduction and basic idea}
\label{}

Many tasks have recently been effectively solved by neural networks.
Deep neural networks are the most actively studied, the intensive progress
in their development is associated with successful solution
of the problem of vanishing and exploding gradients in deep networks. 

In practical use, the problem often arises of constructing from already trained deep network 
a smaller network  allowing one to successfully
solve the problem with  similar  quility\cite{KnowledgeDistilation}.
The main theoretical basis for the possibility of creating small networks
are the theorems of Kolmogorov-Arnold \cite{Kolmogorov_1957,Arnold_1963},
Cybenko \cite{Cybenko_1989}, Funahashi \cite{FUNAHASHI_1989}, and
Hornik \cite{HORNIK_1989,HORNIK_1991} (also known as the universal
approximation theorem) .

In accordance with the Kolmogorov-Arnold theorem, the approximation
of an unknown function of N variables can be represented as:

\begin{equation}
f(\overrightarrow{x})=\sum_{q=0}^{2N}\Phi_{q}\left(\sum_{p=1}^{N}\phi_{q,p}\left(x_{p}\right)\right)\label{eq:KolmoArnold}
\end{equation}

The theorem is not constructive, and proves the existence
of optimal functions of one argument $\Phi_{q}\left(x\right),\phi_{q,p}\left(x\right)$
that provide such an approximation, but does not specify their specific
shape.

The theorems \cite{Cybenko_1989,FUNAHASHI_1989,HORNIK_1989,HORNIK_1991,Sonoda_2017}
specify the types of these functions and prove the possibility
of using various functions $\Phi_{q}\left(x\right),\phi_{q,p}\left(x\right)$. 
Thus, theoretically,
most of the practical problems should be solved by a neural network
with one hidden layer by choosing correct functions $\Phi_{q}\left(x\right),\phi_{q,p}\left(x\right)$.

Traditionally, however, for large input vectors this hidden layer size and
number of unknown functions becomes as huge as $O(N^2)$, where N is dimension of input vector. 
Therefore the solution of many problems
is reduced to the construction of neural networks with a larger number
of hidden layers, and smaller number of neurons in each layer ('thin networks'), 
that efficiently decrease number of unknown (trained) parameters. 
However, there are still problems in which the use
of a large number of hidden layers is not required, and the problem
can be solved with a simpler network.

The simplest
form of the function $\phi_{q,p}$ is linear:

\begin{equation}
\phi_{q,p}(x)=A_{p,q}x+B_{p,q}\label{eq:lin_func}
\end{equation}

and the $\Phi_{q}$ function is the neuron activation function. This
allows one to build neural networks with one hidden layer to solve
some problems 
by choosing corresponding activation function and fitting coefficients $A_{p,q},B_{p,q}$.

A difficult task is to choose the function $\Phi_{q}$ - the 
activation function of the hidden layer. The theorem
\cite{HORNIK_1991} proves that as this function one can use a sufficiently
arbitrary function - continuous and bounded. However even non-bounded functions can be used, 
for example widely used ReLU \cite{ReLU}, that satisfy less strong conditions 
\cite{Sonoda_2017} and also could be used in universal approximators(networks).

The only problem is to find the coefficients $A_{p,q},B_{p,q}$. In modern time this problem 
is solved by gradient descent methods.
Modern experience in building deep neural networks shows that in
order to avoid vanishing and exploding gradients \cite{Bengio_1994}
when searching for optimal network coefficients using the gradient descent
method, only some activation functions are beneficial. This greatly
limits the activation functions suitable for building deep neural
networks. For optimal activation functions, the derivative of the
function should be as close as possible to 1. This is one of the reasons
for replacing Sigmoid (logistic function) with Tanh (hyperbolic tangent)
in the LeNet\cite{Lenet5} network, replacing Tanh with ReLU (Rectified
Linear Unit) in the AlexNet\cite{AlexNet} network, and the emergence
of Residual blocks and ResNet\cite{ResNet}
networks, that are widely used today. 

Therefore, the requirement that the derivative of the activation function
be close to 1 is a very important requirement when choosing activation
functions for deep neural networks. An ideal function that satisfies
this condition is linear. This function is not useful, since it will not
allow effective building deep networks - the superposition of layers 
with linear activation functions
is equivalent to a single layer. Thus, we face with a contradiction,
which is usually resolved by using the ReLU and similar right-hand-linear
functions (SeLU, SWISH, MISH, APTx, etc.) or Residual blocks.

However, this contradiction is apparent: in order for the gradient
to do not vanish or explode, a slightly different condition should be met. Namely,
not the derivative, but the modulus of the derivative of the activation
function must be equal to 1. This property is possessed not only by
a linear function, but also by many other real-valued functions whose derivative modulus
are equal to 1 almost everywhere. 
We can classify these functions by the number of points where their differential is not defined.
The simplest function with none of these points is linear, and the function with single point 
is absolute value (module).
So we can use absolute value function as an activation function (Absolute activation function, Abs),
and $\Phi_{q}$ becomes:

\begin{equation}
\Phi_{q}(y)=W_{q}|y|\label{eq:abs_act_initial}
\end{equation}

It should be noted that most of the popular activation functions (Sigmoid,
Tanh, ReLU, SeLU, LeakyReLU, etc.) are monotonically nondecreasing. Absolute activation function Abs 
differs significantly from them - it is
nonmonotone, but continuous, like SWISH, MISH and APTx functions\cite{APTX}.
The Absolute activation function already mentioned by researchers \cite{AANN_2018,Apicella_2021}, 
but its efficiency has not been demonstrated in details.
The Absolute activation function satisfy conditions of 
\cite{Sonoda_2017} and therefore could be used in universal approximators (neural networks).

A neural network with one hidden layer with an N-dimensional
input vector and Absolute activation will implement the operation:

\begin{equation}
f(\overrightarrow{x})=\sum_{q=0}^{2N}W_{q}\left|\sum_{p=1}^{N}\left(A_{p,q}x_{p}+B_{p,q}\right)\right|\label{eq:abs_activation}
\end{equation}

It can be seen from the eq.(\ref{eq:abs_activation}) that the
network is implemented by two layers of neurons - the hidden
fully connected layer with an Absolute activation function (Abs,
$\varphi(y)=|y|$) and with the number of neurons 2N+1, where N is
the dimension of the input vector. The decision layer is standard
for the given problem: with a linear activation function in the case
of solving the approximation problem, as in (\ref{eq:abs_activation})
or with a Softmax function in the case of solving the classification
problem:

\begin{equation}
f_{c}(\overrightarrow{x})=Softmax_{c}\left(\sum_{q=0}^{2N}W_{q,c}\left|\sum_{p=1}^{N}\left(A_{p,q}x_{p}+B_{p,q}\right)\right|\right)\label{eq:abs_activation-1}
\end{equation}

where c is the class number and $W_{q}$ are the coefficients of the
decision layer.

\section{Simplest 2D classification problems}

\begin{figure}
\includegraphics[scale=0.55]{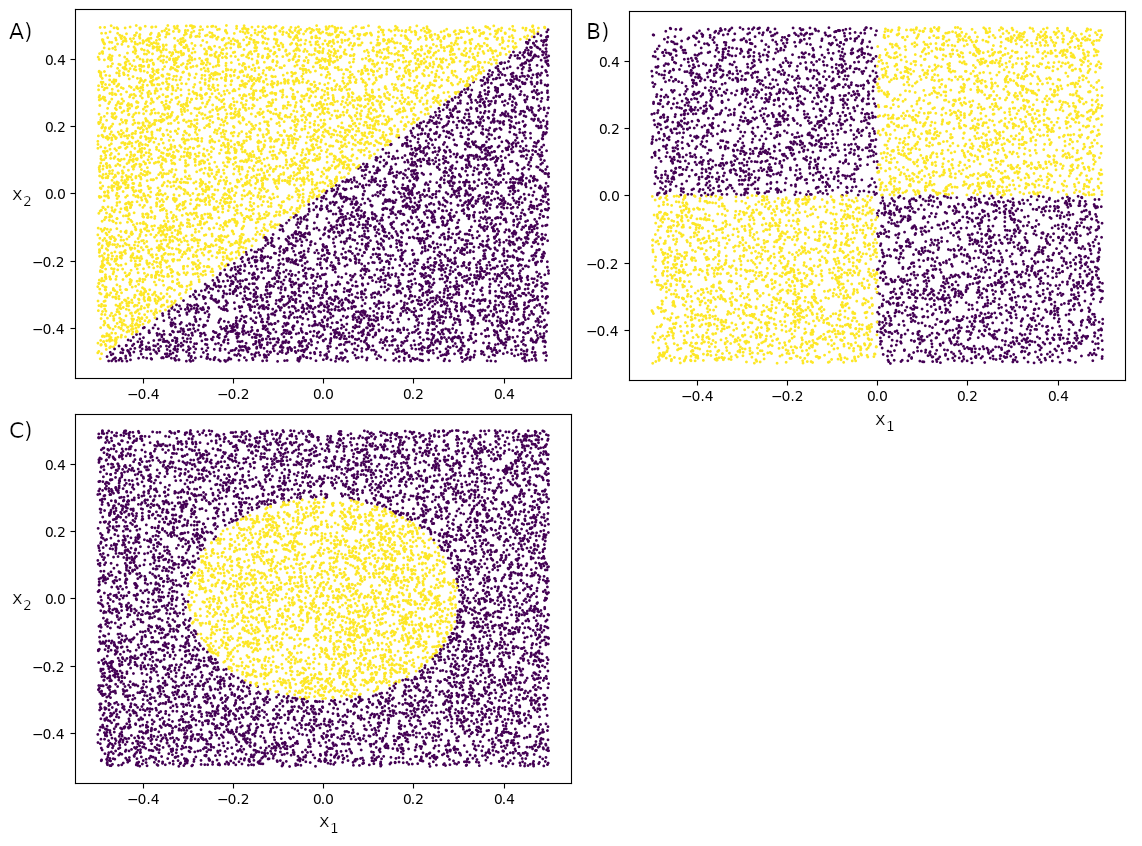}
\caption{Synthetic datasets for the simplest classification: A) linear separation;
B) “cross” separation; C) circular area}
\label{fig:simple_synth}
\end{figure}

Let us demonstrate the possibility of using the Absolute activation
function for effective solving simple classification problems. Let us compare
three networks: a fully-connected network with single hidden
layer with ReLU activations, a fully-connected network with two
hidden layers with ReLU activations, and a fully-connected network with single
hidden layer with Abs activations. We will use two-dimensional
input vectors (N=2). 
The three synthetic problems of two-class classification were solved: a) linear
separation, b) “cross” type separation, c) circular area separation. These
tasks are shown in Fig.\ref{fig:simple_synth}.

In accordance with (\ref{eq:abs_activation})
and the Kolmogorov-Arnold theorem (\ref{eq:KolmoArnold}), we choose
the number of neurons in each layer equal to 2N+1=5. The number of
training epochs is 1000, loss function is cross-entropy, the optimum search method 
is ADAM \cite{ADAM} with learning rate $10^{-3}$. The first network has two hidden layers and
57 unknown parameters, the second and third networks have one hidden
layer and 27 unknown parameters.

\begin{figure}
\includegraphics[scale=0.55]{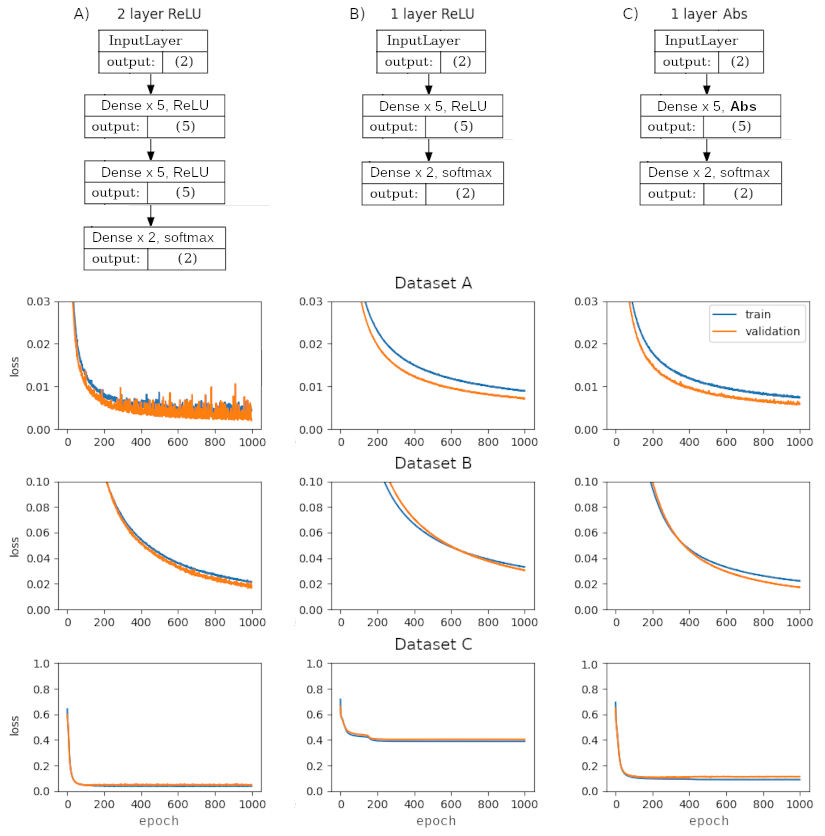}
\caption{The architecture of the studied simple networks: a) two hidden layers
with ReLU activations, b) single hidden layer with ReLU activations c)
single hidden layer with Abs activations. Results of training three networks
on datasets (A-C, as shown in Fig.\ref{fig:simple_synth}). For each
dataset, the dependence of the loss function on the training epoch
on the training and validation datasets is shown.}
\label{fig:simple_d1-3}
\end{figure}

 The architectures
of the studied networks and training processes are shown in Fig.\ref{fig:simple_d1-3}.
Fig.\ref{fig:simple_d1-3} also shows the results of training the networks
on these synthetic test datasets (A-C).

It should be noted that the solution of datasets (B-C) by a network
with single hidden ReLU layer looks unstable. In the most cases it predicts
accurate classification, but in some cases - an inaccurate one 
(see Fig.\ref{fig:simple_d1-3}B for datasets B and C).
Experiments shows, that for dataset B inaccurate training (final accuracy is less than 0.8) is about 9\% of all cases,
for dataset C inaccurate training is about 16\% of all cases. 
For two hidden layer ReLU network inaccurate training probability is about 3\% of B-C cases 
(possibly due to small number of training epochs), 
for single hidden layer Abs network there are no inaccurate runs.

This demonstrates a dependence of the single fully-connected hidden layer ReLU network 
on the initial conditions (initial coefficients values), 
and weaker dependence of the single fully-connected hidden layer Abs network results on initial conditions.

Fig.\ref{fig:simple_d1-3} shows that the neural network with single hidden
layer and Abs activations is no less efficient than the network with
single ReLU hidden layer, but less efficient than the network with two
ReLU hidden layers. At the same time, it has the same number of free
parameters as the network with one hidden ReLU layer and is less dependent
on initial conditions during training. So from the first view Abs activation function 
can be effectively used in classification networks and may improve their efficiency.
Let us study the use of Absolute activation in more complex classification problems.

\section{LeNet-kind solution of MNIST problem}

\subsection{Preliminary analysis}

 Let us consider the
solution of the MNIST problem - the task of classifying handwritten
digital characters into 10 classes\cite{BestMNIST}.

When solving the MNIST problem, the LeNet-5\cite{Lenet5} network
was frequently used as a baseline solution, which is quite compact and fast. 
To demonstrate the effectiveness of Absolute
activation, the network was analyzed in two versions - in its standard
variant (with Tanh activation functions, LeNet-5) and its modernization, 
obtained from LeNet-5 by replacing all the activation functions by Abs (referred as LeNet+Abs).

\begin{figure}
\includegraphics[scale=0.45]{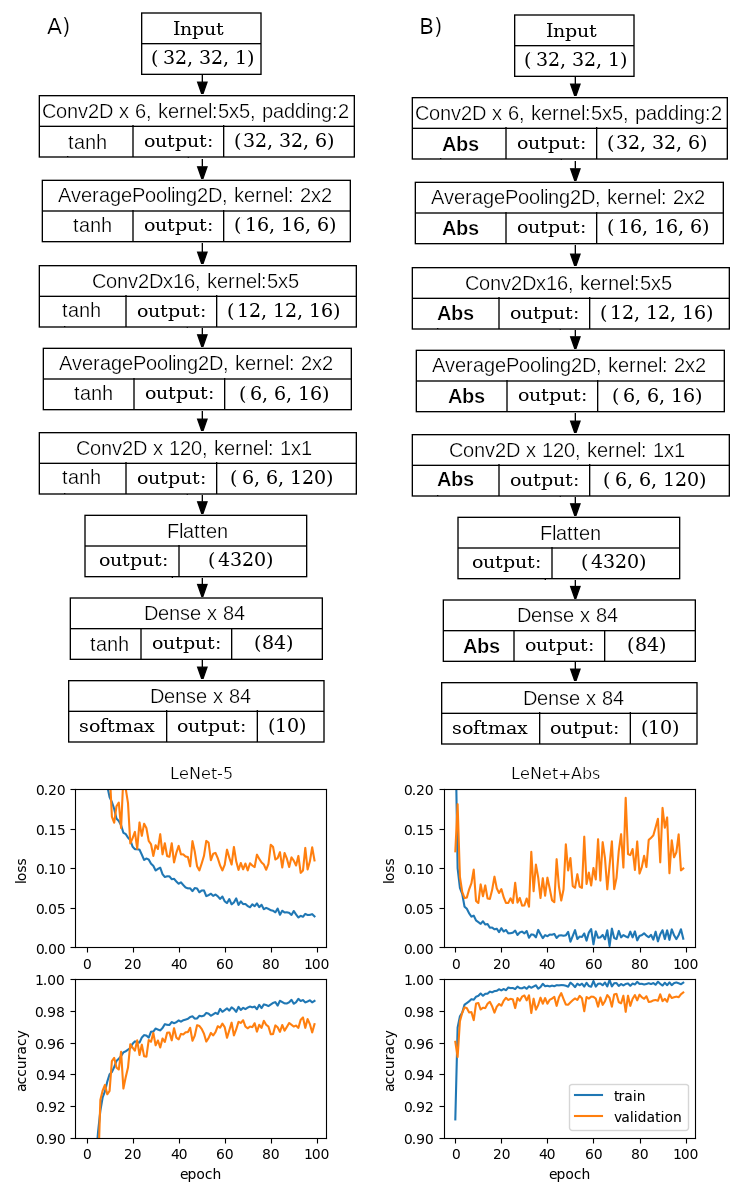}
\caption{Tested network architectures: A) standard LeNet-5; B) LeNet+Abs (with
Absolute activation functions). The results of training the two networks,
from top to bottom: loss on the test and validation datasets, accuracy
(accuracy) on the training and validation datasets.}
\label{fig:res_big_nn}
\end{figure}

Both networks have 368,426 free parameters, network architectures
are similar and shown in Fig.\ref{fig:res_big_nn}. Significant differences are highlighted
in bold. There are also more efficient networks exist for solving the MNIST
problem, for example listed in \cite{Apicella_2021}, but usualy they contain a
larger number of free parameters (for example as much as 2,888,000, as in \cite{BestMNIST}).
Therefore, to demonstrate the effectiveness of the Absolute activation
function, we will use a relatively simple LeNet-5 as baseline solution.

Fig.\ref{fig:res_big_nn} shows the dependencies of the loss function
and accuracy of both networks
after 100 training epochs on identical MNIST training and validation datasets 
(80\% training data followed by 20\% of validation data). 
Optimization algorithm is ADAM with batchsize 128 and learning rate $10^{-3}$.

It can be seen from Fig.\ref{fig:res_big_nn} that the resulting accuracy
of the LeNet+Abs network looks higher than the accuracy of the standard
LeNet-5, and with an accurate training stop, it should provide the better accuracy and
the smaller value of the loss function on the validation dataset, than standard LeNet-5 does.
Thus, using the Abs activation function in this problem at first
view is more profitable than the use of Tanh activations due to
the greater accuracy achieved.

\subsection{Training process details}

It can be seen from the Fig.\ref{fig:res_big_nn} that the dependence
of the LeNet+Abs accuracy and loss function on the training epoch
is not smooth, but much more volatile than those for LeNet-5. There
seems to be two reasons for this. First, when training the network,
by default, the epoch accuracy is calculated over the last batch, i.e.
not over the entire training dataset. The second is that the derivative
of the Absolute activation function, indicating the direction of the
gradient, is discontinuous at zero argument, so small changes in the argument
can lead to sharp changes in the gradient, so the ADAM algorithm
does not provide a smooth enough descent, 
and one should use lower learning rates, slowing down training process.

To find the optimally trained model 
we should find the epoch with maximal accuracy at unknown test dataset.
We do not use the test dataset at training, so we may only estimate this accuracy.
To prevent overtraining, we should not use train dataset information, and
only validation dataset could be used.

Optimally trained network should provide good accuracy exceeding a given limit at almost any test dataset.
After the network trained this minimal accuracy bound should be as high as possible.

To estimate the lower bound of accuracy at any test dataset we should know the distribution 
of network accuracy at different validation datasets. 
We have only single validation dataset, so the lower bound could be estimated
by bootstrap algorithm \cite{Bootstrap}.
But for using bootstrap we should accurately choose significance level, 
that defines how low the lower bound over test dataset should be below the mean validation accuracy.
As shows experiments, more accurate results can be obtained when we just choose minimal 
accuracy over two halves of validation dataset:

\begin{equation}
ACC_{expected@test}=\min (ACC(ValidationData1),ACC(ValidationData2))
\label{eq:acc1}
\end{equation}

Another training problem is choosing a right learning rate for training process to be fast and accurate.
We reach this by by step-by-step decrease learning rate with reinitializaing ADAM between steps.
The ADAM algorithm is known to be very affective, but have hidden parameters, optimized during training.
Reinitialization resets these hidden parameters to their initial values.
Training process at given learning rate stops if expected accuracy lower bound has not increased for 
10 epoches since epoch of previous increase.
During training we use learning rates stepping from $10^{-3}$ downto $10^{-6}$ with 10 times decrease 
between steps. This approach is close to well known ReduceLROnPlateau algorithm with patience 10, 
monitoring of $ACC_{expected@test}$ and reinitializing ADAM before reduction.

The training algorithm is shown as Algorithm \ref{alg:1}

\begin{algorithmic}
\begin{algorithm}
\caption{Training procedure}

\Procedure {CallAfterEachTrainingEpoch}{$Model, EpochNumber, ValidationDataset$}
\State $ExpectedAccuracy \gets ACC_{expected@test}(ValidationDataset)$
\If{$ExpectedAccuracy > BestAccuracy$}
   \State $BestAccuracy \gets ExpectedAccuracy$
   \State $BestEpoch \gets EpochNumber$
   \State SaveModel(Model)
\EndIf
\If{$EpochNumber> BestEpoch+10$}
 \State StopTrainingModel()
\EndIf
\EndProcedure

\For{$LR \in [10^{-3} .. 10^{-6}, step: \times 0.1]$}
\State $ADAM \gets InitializeADAM(LearningRate=LR)$
\If{$LR < 10^{-3}$}
   \State $Model \gets RestoreModelFromSaved()$
\Else 
   \State $Model \gets InitModel()$
\EndIf
\State $BestEpoch \gets 0, BestAccuracy \gets 0$
\State $TrainModelUntilStop(Model,ADAM,TrainDataset,ValidationDataset)$
\EndFor
\label{alg:1}
\end{algorithm}
\end{algorithmic}

The Table \ref{tab:lenet-deep-results} shows the accuracy achieved and total epoch number during training of the
standard LeNet-5 model and the LeNet+Abs model (both marked as "Not degraded" in the Table \ref{tab:lenet-deep-results}). 
The table shows that the accuracy of LeNet+Abs exceeds LeNet-5 accuracy, and the number of errors
on the test dataset has fallen from 1.14\% (for LeNet-5) 
to 0.56\% (for LeNet+Abs) - i.e. almost 2 times. 
Also the found error level 0.56\% for LeNet+Abs looks better than the lowest error level 0.7\% reported in \cite{Lenet5} 
even for boosted variants of LeNet.
One can also see that training of LeNet+Abs slower than training LeNet-5 for about 1.5 times (100 training epochs vs. 68 
correspondingly). 

\subsection{Gradient vanishing robustness}

It is known that an important problem in training deep and recurrent
networks is gradient vanishing and explosion - the loss of the network's ability to
train layers with an increase of their number \cite{Bengio_1994}.
Recently, Residual blocks\cite{ResNet} have been found to solve this
problem. Let us demonstrate that the Absolute activation function
is resistant to the gradient vanishing and explosion effect, 
not worse than popular ReLU and SeLU functions. 

\begin{figure}
\includegraphics[scale=0.3]{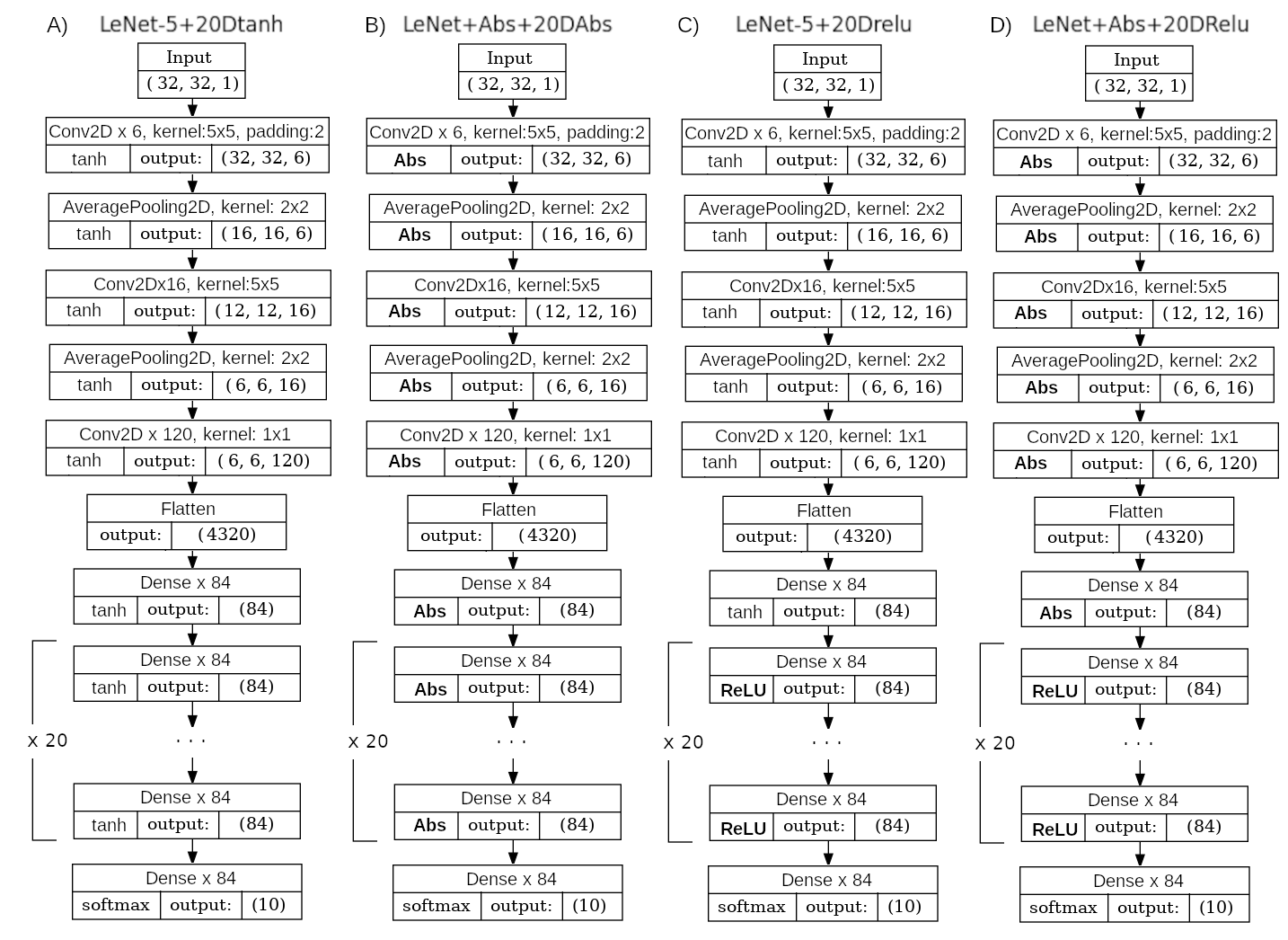}
\caption{Degraded versions of LeNet-5 and LeNet+Abs networks to test the effect
of gradient vanisning and explosion effect: LeNet-5+20DTanh (A), LeNet+Abs+20DAbs
(B) , LeNet-5+20Drelu (C), LeNet+Abs+20Drelu (D). Architecture and
quality of training.}
\label{fig:lenet_megadeep}
\end{figure}

To show this, inside each of the networks (LeNet-5 and LeNet+Abs) 20 intermediate
(disturbing) layers were placed (as the last hidden layers), complicating
training the network by vanishing gradient. To LeNet-5 and LeNet+Abs architectures a 20 layers 
with Tanh, ReLU, SeSU, Abs activations were added, producing 8 final architures:
Lenet-5+20DTanh, Lenet-5+20DReLU, Lenet-5+20DSeLU, Lenet-5+20DAbs,
Lenet+Abs+20DTanh, Lenet+Abs+20DReLU, Lenet+Abs+20DSeLU, Lenet+Abs+20DAbs. 
Examples of these architectures are shown in Fig.\ref{fig:lenet_megadeep}.

All 8 resulting (deep) networks have 511226 free coefficients, and about 143
thousand of them are from the last (disturbing) layers, and these networks
differ only by activation functions in layers.

The training results are shown in Table \ref{tab:lenet-deep-results}.

It can be seen from the Table  \ref{tab:lenet-deep-results} that adding 20 more hidden 
layers with Abs activation at the end makes it still possible to train LeNet+Abs+20DAbs,
although with somewhat lower accuracy  than in
the absence of these layers - LeNet+Abs network (99.05\% vs. 99.44\% correspondingly). 
This confirms with
our theoretical expectation that the Absolute activation function
does not cause significant gradient vanishing or explosion, and therefore
can be effectively used in both simple and deep neural networks.
It can be seen from Table \ref{tab:lenet-deep-results} that the gradient vanishing/explosion it causes is better
than that of ReLU, and that Abs as good as SeLU activation. This allows to use Abs in deep networks.

\begin{table}
\caption{Accuracy (\%) of various networks - standard LeNet-5 and LeNet+Abs, 
degraded by the last 20 hidden fully connected layers with different activation functions. 
*) marks the cases when learning rate $10^{-3}$ is not enought to start training 
and we start from $10^{-4}$ instead. Bootstrap confidential interval over test dataset at standard confidence level 0.95 
is also shown.}
\centering

\begin{tabular}{|c|c|c|c|}
\hline 
Network & Accuracy & Trained Epoches & Conf.interval\tabularnewline
\hline 
\hline 
\multicolumn{4}{|l|}{LeNet-5 architecture}\tabularnewline
\hline 
\hline 
Not degraded & 98.86 & 68 & (98.65, 99.07) \tabularnewline
\hline 
+20DTanh & 98.41 & 80 & (98.17,98.66) \tabularnewline
\hline 
+20DReLU & 98.32 & 129 & (98.07,98.57) \tabularnewline
\hline 
+20DSeLU & 98.75 & 80 & (98.53,98.97) \tabularnewline
\hline 
+20DAbs(*) & 98.71 & 66 & (98.49,98.93) \tabularnewline
\hline 
\hline 
\multicolumn{4}{|l|}{LeNet+Abs architecture}\tabularnewline
\hline 
\bf{Not degraded} & \bf{99.44} & 100 & (99.30,99.59) \tabularnewline
\hline 
+20DTanh(*) & 97.51 & 80 & (96.90,97.54) \tabularnewline
\hline 
+20DReLU & 98.92 & 118 & (98.72,99.12) \tabularnewline
\hline 
+20DSeLU & 99.03 & 98 & (98.84,99.22) \tabularnewline
\hline 
\bf{+20DAbs} & \bf{99.05} & 112 & (98.86,99.24) \tabularnewline
\hline 
\end{tabular}
\label{tab:lenet-deep-results}
\end{table}

\subsection{Reducing the LeNet-5 network size without loss of accuracy}

\begin{figure}
\includegraphics[scale=0.35]{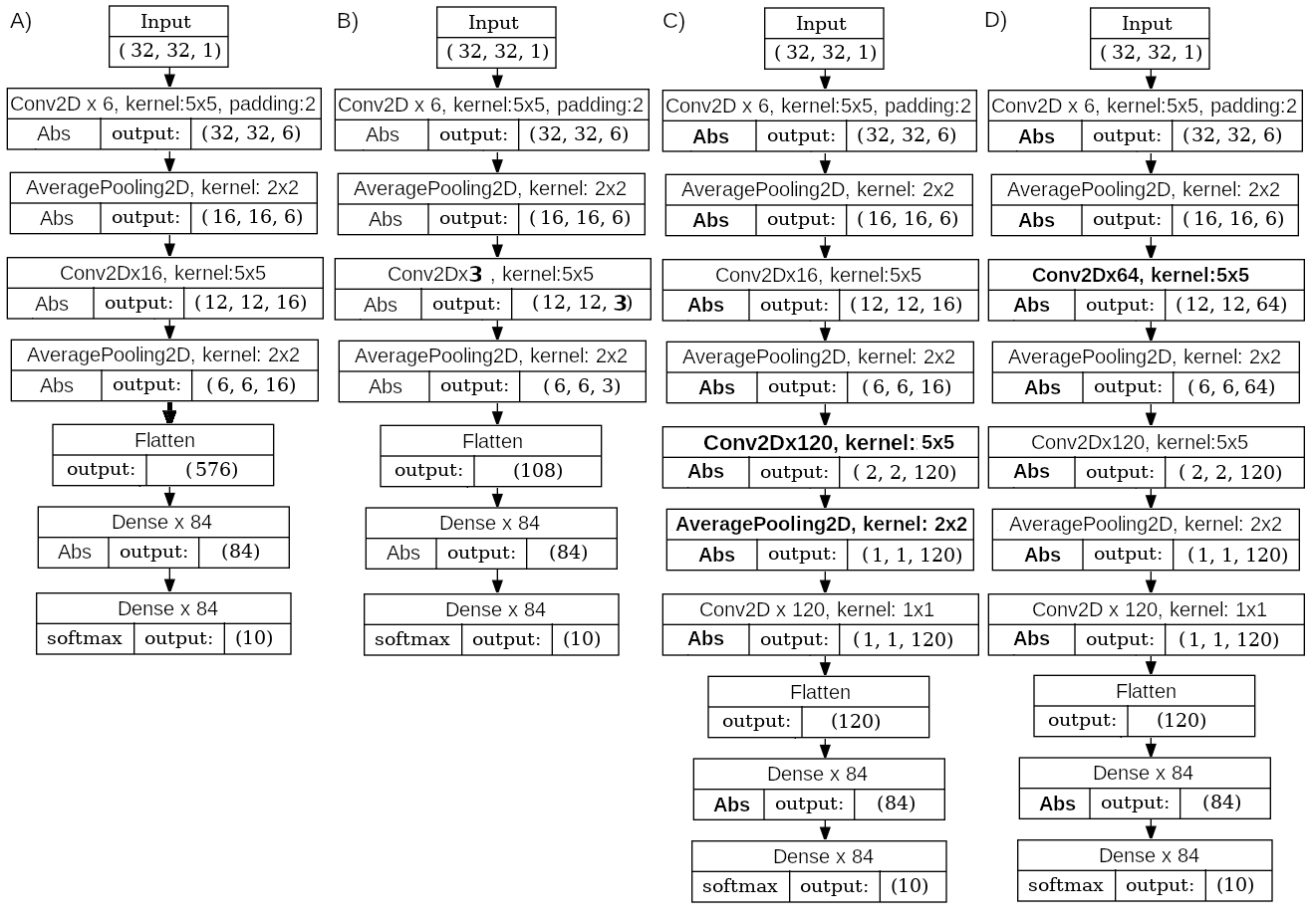}
\caption{Smaller models A)SmallLeNet+Abs; B)TinyLeNet+Abs; C)LeNet+Conv120+Abs; D)LeNet+Conv64+Conv120+Abs.
Bold highlights the main difference from the previous model.}
\label{fig:small_tiny_ultratiny}
\end{figure}

The good stability of the Absolute activation function against gradient
vanishing suggests that the original LeNet-5 neural network could be too complex
for its reported accuracy. Let us reduce its size without loosing the accuracy. 
The most
obvious way is to remove the last hidden convolutional layer from
LeNet+Abs. This resulting network referred as SmallLeNet+Abs network
and shown in Fig.\ref{fig:small_tiny_ultratiny}A.
The TinyLeNet+Abs model is made from the SmallLeNet+Abs model by
reducing the number of convolutions in the second convolution layer
from 16 to 3. This new architecture
is also shown in Fig.\ref{fig:small_tiny_ultratiny}B. 
Bold marks important changes.

The Table \ref{tab:acc_lenet-1} shows the accuracies achieved by these networks using 
training Algorithm \ref{alg:1}. In addition, the number
of trained model coefficients, total trained epoches, 
and bootstrap confidence interval on
the test dataset are shown (with a standard significance level of
0.95).

The Table \ref{tab:acc_lenet-1} shows that the best accuracy on the
test dataset is provided by the LeNet+Abs model.
The SmallLeNet+Abs model with 52 thousand trained parameters,
and TinyLeNet+Abs model with 10.6 thousand trained parameters 
also exceeds the initial LeNet-5 (with 368 thousand parameters) accuracy in most cases.

More accurate network is extended LeNet architecture - Lenet+Conv120+Abs, created  
by adding Conv120 and AveragePooling layers to LeNet (it has 76226 parameters
and is shown in Fig.\ref{fig:small_tiny_ultratiny}C), 
provides the even better accuracy than LeNet+Abs architecture - 99.51\%, but having 4.8 times less parameters.

Thus, the replacement of Tanh activation functions with
Abs makes it possible to reduce the size of original LeNet-5
model by more than 4.8 and 7 times with an improvement of its accuracy 
(LeNet+Conv120+Abs and SmallLeNet+Abs variants correspondingly), 
and providing better accuracy than ReLU and SeLU variants. 
It can be seen from the bootstrap confidence intervals that even TinyLeNet+Abs
model with the number of parameters 35 times less than the number
of LeNet-5 parameters provides an accuracy that not worse than the accuracy
of original LeNet-5 with Tanh, ReLU, or SeLU activations.

\begin{table}
\caption{The best accuracy obtained on test dataset for 
different architectures with different activation functions.
The number of trained model parameters and the bootstrap
confidence interval (standard confidence level 0.95) over the test dataset. Bold marks best activation 
function for given architecture. $ACC_{expected@test}$ according to eq.\ref{eq:acc1}. }
\centering
\begin{tabular}{|p{2cm}|c|c|p{2.9cm}|p{1.5cm}|}
\hline 
 & Accuracy & Trained epoches & Conf.interval & \# coefs \tabularnewline
\hline 
\hline 
\multicolumn{4}{|l|}{LeNet architecture} & 368426\tabularnewline
\hline 
+Tanh (LeNet-5) & 98.86 & 68 & (98.65, 99.07) & 368426\tabularnewline
\hline 
+ ReLU & 99.13 & 72 & (98.95,99.31) & -\tabularnewline
\hline 
+ SeLU & 99.03 & 62 & (98.84,99.23) & -\tabularnewline
\hline 
\bf{+ Abs} & \bf{99.44} & 100 & (99.30,99.58) & -\tabularnewline
\hline 
\hline 
\multicolumn{4}{|l|}{SmallLeNet architecture} & 51890\tabularnewline
\hline 
+ Tanh & 99.03 & 64 & (98.84, 99.22) & 51890\tabularnewline
\hline 
+ Relu & 99.10 & 88 & (98.91, 99.28) & -\tabularnewline
\hline 
+ Selu & 99.10 & 59 & (98.92, 99.28) & -\tabularnewline
\hline 
\bf{+ Abs} & \bf{99.28} & 116 & (99.12, 99.44) & -\tabularnewline
\hline 
\hline 
\multicolumn{4}{|l|}{TinyLeNet architecture} & 10615\tabularnewline
\hline 
+ Tanh & 98.76 & 75 & (98.54, 98.97) & 10615\tabularnewline
\hline 
+ Relu & 98.94 & 71 & (98.74, 99.14) & -\tabularnewline
\hline 
+ Selu & 98.65 & 74 & (98.42, 98.88) & -\tabularnewline
\hline 
\bf{+ Abs} & \bf{99.08} & 87 & (98.90, 99.27) & -\tabularnewline
\hline 
\hline 
\multicolumn{4}{|l|}{LeNet+Conv120 architecture} &76226\tabularnewline
\hline 
+ Tanh & 99.09 & 72 & (98.91, 99.27) & -\tabularnewline
\hline 
+ Relu  & 99.28 & 80 & (99.12, 99.45) & -\tabularnewline
\hline 
+ Selu  & 99.26 & 65 & (99.09, 99.43) & -\tabularnewline
\hline 
\bf{+ Abs}  & \bf{99.51} & 103 & (99.38, 99.65) & -\tabularnewline
\hline 
\hline 

\end{tabular}
\label{tab:acc_lenet-1}
\end{table}

As we have shown, even standard LeNet-5 with Abs activation (LeNet+Abs) can 
provide very good accuracy (99.44\%)
comparable with top solutions for fixed shape activation functions \cite{Apicella_2021}.
Extended LeNet versions (Lenet+Conv120+Abs) have 
accuracy up to  (99.51\%), very close to top solutions for fixed shape activation 
functions \cite{Apicella_2021}, and having less than 77,000 trained parameters.

\section{More efficient accuracy predictions for training}

As we have shown above, the networks with Abs activation looks better than the networks with ReLU, 
SeLU and Tanh activations.
But there are two problems arise.

 Problem 1. The stabuility of training the network. We use random initial network parameters, 
no regularization, so the found loss function minimum is not global one. We need to check how the 
initializing and training affect the trained network accuracy.

 Problem 2. The stop condition. As a stop condition we use prediction of network accuracy at unknown 
test dataset over validation dataset. It is obvious that different prediction algorithms lead to different network accuracy.

To study these problems we created 3 ensembles for LeNet+Conv120 network architecture (LeNet+Conv120+Abs,
LeNet+Conv120+ReLU, LeNet+Conv120+SeLU, LeNet+Conv120+Tanh), that differs only by activation function (Abs, ReLU, SeLU, Tanh).
Each of 3 ensembles is trained by using different model for accuracy at test dataset prediction 
over validation dataset ( $ACC_{expected@test}$ in Algorithm \ref{alg:1} ):

\begin{equation}
ACC_{expected@test,1}=\min (ACC(ValidationData1),ACC(ValidationData2))
\label{eq:accf1}
\end{equation}

\begin{equation}
ACC_{expected@test,2}=MEAN_{bootstrap}(ValidationData)-STD_{bootstrap}(ValidationData)
\label{eq:accf2}
\end{equation}

\begin{equation}
ACC_{expected@test,3}=ACC_{expected@test,1}-STD_{bootstrap}(ValidationData)
\label{eq:accf3}
\end{equation}

The first prediction method is discussed earlier with Algorithm \ref{alg:1}, the second one is based on bootstrap estimate 
of lower bound of prediction interval over validation dataset with significance level 0.68, 
the last one is combination of the first and the second, lowering the expected bound. 

Each ensemble of trained networks were used to produce 2 values, shown in Table \ref{tab:ensemble_results}.
The first one is the accuracy over the test dataset by majority ensemble voting, when the predicted label is found 
as the label predicted by maximal number of networks (marked as M.V. - majority voting).
Another measure is prediction interval over the test dataset over the ensemble of trained networks 
(marked as C.I. - confidence interval).

For different purposes different M.V. and C.I. are required. 
For example, for majority ensemble voting the networks should be overtrained 
and M.V. should 
be as high as possible. As shown in Table \ref{tab:ensemble_results}, for Abs network the best M.V. is 
provided by use $ACC_{expected@test,1}$ or $ACC_{expected@test,3}$ accuracy prediction variant during train.

When only single network is planned for prediction, we need the highest confidential interval.
In this case using $ACC_{expected@test,3}$ prediction variant is preferable.
For both cases using $ACC_{expected@test,3}$ accuracy prediction for training the LeNet+Conv120+Abs is preferable.

To improve the result, a  LeNet+Conv64+Conv120+Abs were made - it is wider and deeper LeNet-like network,
having more convolutions at top layer, than  LeNet+Conv120+Abs has.
Its architecture is shown in Fig.\ref{fig:small_tiny_ultratiny}D. 

The model has 227474 parameters, that is also smaller than number of parameters in original LeNet-5 network, 
and provides better accuracy than LeNet+Conv120+Abs.
It's ensemble training using  $ACC_{expected@test,3}$ test dataset accuracy prediction shows its increased accuracy, 
also exceding its ReLU, SeLU and Tanh varaints (shown in Table \ref{tab:ensemble_results}).

\begin{table}
\caption {
Accuracy results for ensemble of from 20 training runs, produced 20 trained variants of LeNet+Conv120 and LeNet+Conv64+Conv120 networks,
with different algorithms for predicting accuracy at test dataset from the validation dataset during training, 
eqs.\ref{eq:accf1}-\ref{eq:accf3}.
M.V. rows - majority voting ensemble accuracy over 20 networks at test dataset,
C.I. rows - accuracy limits over the ensemble at test dataset. Best results marked by bold.}
\centering

\begin{tabular}{|l|c|c|c|}
\hline 
  & $ACC_{expected@test,1}$ & $ACC_{expected@test,2}$ & $ACC_{expected@test,3}$ \tabularnewline
\hline 
\hline 
\multicolumn{4}{|l|}{Lenet+Conv120 architecture} \tabularnewline
\hline 
+Tanh  C.I. & [98.95,99.20] & [98.93,99.21] &  [99.03,99.22] \tabularnewline
\hline 
+Tanh M.V. & 99.33 & 99.32 & 99.36 \tabularnewline
\hline 
\hline 
+ReLU  C.I. & [99.12,99.38] & [99.16,99.37] &  [99.13,99.37] \tabularnewline
\hline 
+ReLU M.V. & 99.52 & 99.52 & 99.48 \tabularnewline
\hline 
\hline 
+SeLU  C.I. & [99.19,99.36] & [99.17,99.39] &  [99.20,99.35] \tabularnewline
\hline 
+SeLU M.V. & 99.54 & 99.55 & 99.54 \tabularnewline
\hline 
\hline 
+Abs  C.I. & [99.39,99.56] & [99.35,99.57] &  \bf{[99.41,99.59]} \tabularnewline
\hline 
+Abs  M.V. & \bf{99.64} & 99.61 & \bf{99.64} \tabularnewline
\hline 
\hline 
\multicolumn{4}{|l|}{Lenet+Conv64+Conv120 architecture} \tabularnewline
\hline 
\hline 
+Tanh  C.I. &  &  &  [98.87,99.15] \tabularnewline
\hline 
+Tanh M.V. & &  & 99.37 \tabularnewline
\hline 
\hline 
+ReLU  C.I. &  &  &  [99.24,99.50] \tabularnewline
\hline 
+ReLU M.V. &  &  & 99.55 \tabularnewline
\hline 
\hline 
+SeLU  C.I. &  &  &  [99.29,99.44] \tabularnewline
\hline 
+SeLU M.V. & & & 99.56 \tabularnewline
\hline 
\hline 
+Abs  C.I. & & &  \bf{[99.45,99.60]} \tabularnewline
\hline 
+Abs  M.V. &  &  & \bf{99.64} \tabularnewline
\hline 
\hline 
\end{tabular}
\label{tab:ensemble_results}
\end{table}

\section{Conclusion}

In the paper we discuss the Absolute activation function for classification neural networks.
An examples of using this activation function in simple
and complex classifying problems are presented. In solving the
MNIST problem with the LeNet-5 network, the efficiency of Abs is shown in
comparison with Tanh, ReLU and SeLU activations. It allows to reach 99.44\% accuracy at standard LeNet-5 network 
by only changing all activations into Abs.
It is shown that its use practically does not lead to gradient vanishing/explosion, and 
therefore Absolute activation can be used in both small and deep neural networks. 
It is shown that in solving the MNIST problem with the LeNet-5 architecture,
the use of Absolute activation helps to significantly reduce the size
of the neural network (up to 2 orders by number of trained parameters)
and improve the accuracy of the solution.
The mean reached accuracy could be about 99.51\% ([99.41\%..99.59\%] at different trained varaints) 
at train dataset with decreasing the network size 
from 368 thousands to 76 thousands coeficients(LeNet+Conv120+Abs) 
and about 99.53\% ([99.45\%..99.60\%] at different trained varaints) at train dataset with decreasing the network size 
from 368 thousands to 227 thousands coeficients(LeNet+Conv64+Conv120+Abs).
Therefore these networks with Absolute activations could be close to the best solutions accuracy 
for fixed shape activation functions, reached by more complicated networks \cite{Apicella_2021}. 

It is demonstrated that training curve (accuracy vs. training epoch) when using Abs activation 
is not smooth, so one should use more complex technique for changing learning ratio and stop training. 
We doing this by estimating the lower bound of network accuracy at any test dataset 
using accuracy calculated over the halves of validating dataset and by bootstrap method and 
combining these predictions.

Thus, the high efficiency of the Absolute activation function has
been demonstrated, which makes it possible to improve the accuracy
of current classification neural networks even without changing their architecture.

\section*{Acknowledgements}
 The data analysis was performed in part on the equipment of the Bioinformatics Shared Access Center, 
 the Federal Research Center Institute of Cytology and Genetics of Siberian Branch of the Russian Academy of Sciences (ICG SB RAS). 
 The work has been done under financial support of the Ministry of Science and Higher Education 
 of the Russian Federation (Subsidy No.075-GZ/C3569/278).


\bibliographystyle{unsrtnat}



\begin{thebibliography}{19}
\providecommand{\natexlab}[1]{#1}
\providecommand{\url}[1]{\texttt{#1}}
\expandafter\ifx\csname urlstyle\endcsname\relax
  \providecommand{\doi}[1]{doi: #1}\else
  \providecommand{\doi}{doi: \begingroup \urlstyle{rm}\Url}\fi

\bibitem[Hinton et~al.(2015)]{KnowledgeDistilation}
Geoffrey Hinton, Oriol Vinyals, and Jeff Dean.
\newblock Distilling the knowledge in a neural network, 2015.

\bibitem[Kolmogorov(1957)]{Kolmogorov_1957}
A.N. Kolmogorov.
\newblock On the representation of continuous functions of many variables by
  superposition of continuous functions of one variable and addition.
\newblock \emph{Dokl. Akad. Nauk SSSR}, pages 953--956, 1957.

\bibitem[Arnold(1963)]{Arnold_1963}
Vladimir Arnold.
\newblock On the function of three variables.
\newblock \emph{Amer. Math. Soc. Transl.}, pages 51--54, 1963.

\bibitem[Cybenko(1989)]{Cybenko_1989}
G.~Cybenko.
\newblock Approximation by superpositions of a sigmoidal function.
\newblock \emph{Mathematics of Control, Signals, and Systems}, 2\penalty0
  (4):\penalty0 303--314, 1989.
\newblock \doi{10.1007/bf02551274}.

\bibitem[Funahashi(1989)]{FUNAHASHI_1989}
Ken-Ichi Funahashi.
\newblock On the approximate realization of continuous mappings by neural
  networks.
\newblock \emph{Neural Networks}, 2\penalty0 (3):\penalty0 183--192, 1989.
\newblock \doi{10.1016/0893-6080(89)90003-8}.

\bibitem[Hornik et~al.(1989)]{HORNIK_1989}
Kurt Hornik, Maxwell Stinchcombe, and Halbert White.
\newblock Multilayer feedforward networks are universal approximators.
\newblock \emph{Neural Networks}, 2\penalty0 (5):\penalty0 359--366, 1989.
\newblock \doi{10.1016/0893-6080(89)90020-8}.

\bibitem[Hornik(1991)]{HORNIK_1991}
Kurt Hornik.
\newblock Approximation capabilities of multilayer feedforward networks.
\newblock \emph{Neural Networks}, 4\penalty0 (2):\penalty0 251--257, 1991.
\newblock \doi{10.1016/0893-6080(91)90009-T}.

\bibitem[Sonoda and Murata(2017)]{Sonoda_2017}
Sho Sonoda and Noboru Murata.
\newblock Neural network with unbounded activation functions is universal
  approximator.
\newblock \emph{Applied and Computational Harmonic Analysis}, 43\penalty0
  (2):\penalty0 233--268, 2017.
\newblock \doi{10.1016/j.acha.2015.12.005}.

\bibitem[Agarap(2018)]{ReLU}
Abien~Fred Agarap.
\newblock {Deep Learning using Rectified Linear Units (ReLU)}, 2018.

\bibitem[Bengio et~al.(1994)]{Bengio_1994}
Y.~Bengio, P.~Simard, and P.~Frasconi.
\newblock Learning long-term dependencies with gradient descent is difficult.
\newblock \emph{{IEEE} Transactions on Neural Networks}, 5\penalty0
  (2):\penalty0 157--166, 1994.
\newblock \doi{10.1109/72.279181}.

\bibitem[Lecun et~al.(1998)]{Lenet5}
Y.~Lecun, L.~Bottou, Y.~Bengio, and P.~Haffner.
\newblock Gradient-based learning applied to document recognition.
\newblock \emph{Proceedings of the IEEE}, 86\penalty0 (11):\penalty0
  2278--2324, 1998.
\newblock \doi{10.1109/5.726791}.

\bibitem[Krizhevsky et~al.(2017)]{AlexNet}
Alex Krizhevsky, Ilya Sutskever, and Geoffrey~E. Hinton.
\newblock {ImageNet} classification with deep convolutional neural networks.
\newblock \emph{Communications of the {ACM}}, 60\penalty0 (6):\penalty0 84--90,
  2017.
\newblock \doi{10.1145/3065386}.

\bibitem[He et~al.(2015)]{ResNet}
Kaiming He, Xiangyu Zhang, Shaoqing Ren, and Jian Sun.
\newblock Deep residual learning for image recognition, 2015.

\bibitem[Kumar(2022)]{APTX}
Ravin Kumar.
\newblock {APTx: better activation function than MISH, SWISH, and ReLU's
  variants used in deep learning}.
\newblock 2022.
\newblock \doi{10.48550/ARXIV.2209.06119}.

\bibitem[Karnewar(2018)]{AANN_2018}
Animesh Karnewar.
\newblock {AANN: Absolute Artificial Neural Network}.
\newblock In \emph{2018 3rd International Conference for Convergence in
  Technology (I2CT)}, 2018.
\newblock \doi{10.1109/i2ct.2018.8529552}.

\bibitem[Apicella et~al.(2021)]{Apicella_2021}
Andrea Apicella, Francesco Donnarumma, Francesco Isgr{\`{o}}, and Roberto
  Prevete.
\newblock A survey on modern trainable activation functions.
\newblock \emph{Neural Networks}, 138:\penalty0 14--32, 2021.
\newblock \doi{10.1016/j.neunet.2021.01.026}.

\bibitem[Kingma and Ba(2014)]{ADAM}
Diederik~P. Kingma and Jimmy Ba.
\newblock {Adam: A Method for Stochastic Optimization}, 2014.

\bibitem[Simard et~al.(2003)]{BestMNIST}
P.Y. Simard, D.~Steinkraus, and J.C. Platt.
\newblock Best practices for convolutional neural networks applied to visual
  document analysis.
\newblock In \emph{Seventh International Conference on Document Analysis and
  Recognition, 2003. Proceedings.} {IEEE} Comput. Soc, 2003.
\newblock \doi{10.1109/icdar.2003.1227801}.

\bibitem[Efron(1979)]{Bootstrap}
B.~Efron.
\newblock {Bootstrap Methods: Another Look at the Jackknife}.
\newblock \emph{The Annals of Statistics}, 7\penalty0 (1):\penalty0 1 -- 26,
  1979.
\newblock \doi{10.1214/aos/1176344552}.

\end{thebibliography}





\end{document}